\documentclass{article} 
\usepackage{iclr2027_conference,times}

\usepackage{amsmath,amsfonts,bm}

\def\eqref#1{equation~\ref{#1}}

\def\1{\bm{1}}

\DeclareMathAlphabet{\mathsfit}{\encodingdefault}{\sfdefault}{m}{sl}
\SetMathAlphabet{\mathsfit}{bold}{\encodingdefault}{\sfdefault}{bx}{n}

\usepackage{hyperref}
\usepackage{url}
\usepackage{graphicx}
\usepackage{booktabs}
\usepackage{subcaption}
\usepackage{todonotes}
\usepackage{listings}
\usepackage{xcolor}

\lstdefinestyle{promptstyle}{
    basicstyle=\ttfamily\scriptsize,
    breaklines=true,
    breakatwhitespace=false,
    columns=fullflexible,
    keepspaces=true,
    frame=single,
    showstringspaces=false
}

\title{JHiMA-WebRL: A Joint Hierarchical Multi-Agent Reinforcement Learning Framework for Long-Horizon Web Tasks
}

\author{Shanka Subhra Mondal \;\; Jiayun Wang \\
Center for Advanced AI, Accenture
}

\iclrfinalcopy 
\begin{document}

\maketitle
\lhead{Preprint}

\begin{abstract}
Multimodal web agents have shown promising capabilities for interacting with websites through graphical user interfaces (GUIs), yet realistic web tasks often require long-horizon, multi-step planning that remains challenging for monolithic policies. Such agents must jointly infer high-level plans, navigate through intermediate states, track progress over extended interaction histories, and execute visually grounded actions. Although prior work has explored modular multi-agent systems with a separate planner, the planner is typically not updated jointly with the actor. We investigate whether explicitly separating and jointly training planning and execution provides a stronger inductive bias for long-horizon web navigation. We introduce JHiMA-WebRL, a joint hierarchical multi-agent reinforcement learning framework for web agents that couples a task decomposer with an actor. To train the decomposer, we use a strong teacher model to construct structured subtask supervision from multimodal GUI trajectories, and use these annotations to fine-tune both the decomposer and the subtask-conditioned actor. We then extend Group Relative Policy Optimization (GRPO) to jointly optimize the decomposer and actor in the interactive WebArena environment through hierarchical rollout grouping and role-specific advantages. On WebArena-Lite-v2, JHiMA-WebRL-14B, consisting of a 7B actor and a 7B decomposer, achieves an average success rate of 37.2\%, compared with 24.0\% for a ScaleCUA-32B agent that we fine-tune with RL on the same training tasks and evaluate under the same protocol. Ablations show that jointly updating the decomposer during RL outperforms updating the actor alone. These results suggest that jointly learned role factorization with smaller models can be a competitive alternative to scaling monolithic agents for long-horizon web tasks.

\end{abstract}

\begin{figure}[h]
    \centering
    \includegraphics[width=\linewidth]{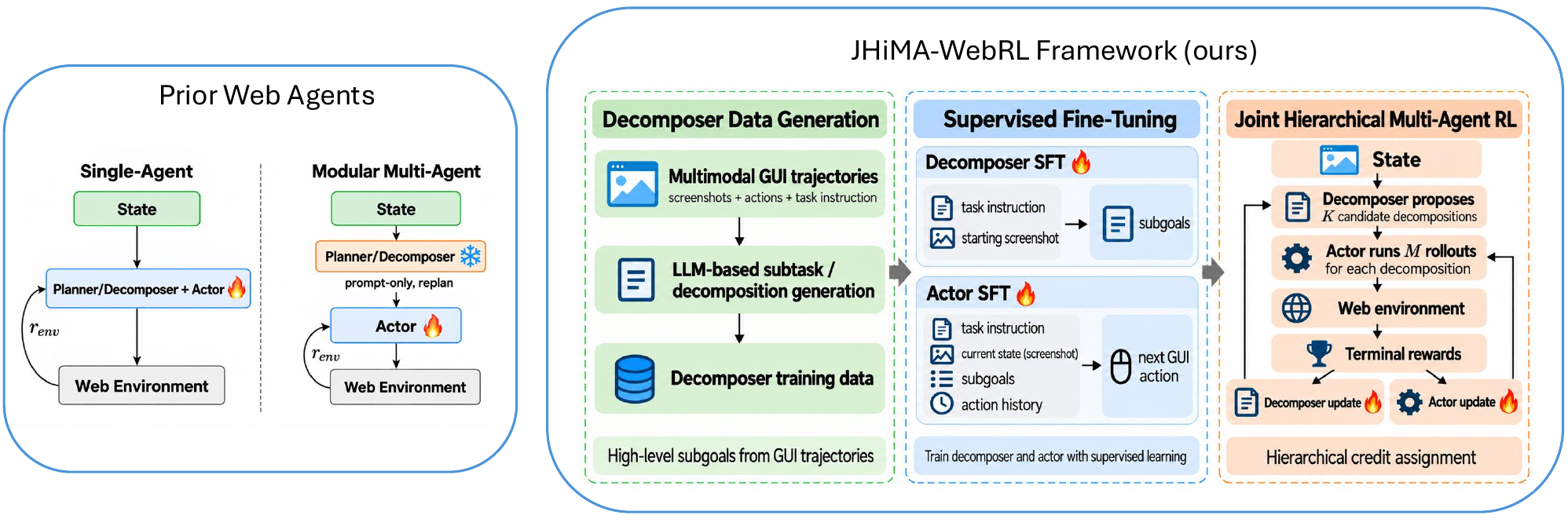}
    \caption{\textbf{Overview of JHiMA-WebRL.} Left: prior web agents use a single policy, or a separate planner that receives no RL updates while the actor is trained. Right: we generate subtask supervision from GUI trajectories, fine-tune the decomposer and actor, and jointly optimize both with hierarchical RL: $K$ decompositions, $M$ actor rollouts each, and role-specific advantages from terminal rewards.}
    \label{fig:multiagent_architecture}
\end{figure}

\section{Introduction}

Multimodal web agents interact with websites by perceiving screenshots, interpreting page elements, and executing actions such as clicking, typing, scrolling, and navigating between webpages. Such agents could support workflows in online shopping, web-based administration, repository management, and information retrieval. Benchmarks such as WebArena \citep{zhou2024webarena} and VisualWebArena \citep{koh2024visualwebarena} evaluate this setting through tasks that require interactive execution across multiple steps. These tasks combine high-level planning with low-level visually grounded control: for example, to cancel an order, an agent must navigate intermediate pages, identify relevant interface elements, track progress over a long action history, and avoid repeated or irrelevant actions.

Many web agents use a monolithic policy that maps the task, observation, and action history directly to the next action, learning planning, state tracking, visual grounding, and execution within a single model. ScaleCUA \citep{liu2025scalecua} and UI-TARS \citep{qin2025uitars} demonstrate the value of large-scale GUI data for training such policies, while WebAgent-R1 \citep{wei2025webagentr1} and WebRL \citep{qi2024webrl} improve web interaction through online reinforcement learning (RL). Modular systems instead introduce a separate planner: Agent S uses a manager to generate subtasks from retrieved knowledge and experience for workers to execute \citep{agashe2025agents}, and Agent S2 adds proactive replanning as execution provides new observations \citep{agashe2025agents2}. Even though an explicit planner module can support long-horizon behavior, the planner's parameters are not updated from the outcomes of action execution. Recent studies \citep{erdogan2025planandact,wu2026plannermatters,kong2026hiroc,lu2025swirl}
focus on training modular agents for long-horizon tasks, but either train the modules separately, apply supervised learning or RL to only one module while freezing the other, or alternate updates between modules. To our knowledge, prior work has not studied simultaneous RL training of a separately parameterized planner and actor from a shared rollout batch in web environments.

We therefore introduce JHiMA-WebRL, a joint hierarchical multi-agent reinforcement learning framework for long-horizon web interaction. Figure~\ref{fig:multiagent_architecture} presents an overview of the three stages in our framework: decomposer data generation, supervised fine-tuning, and joint hierarchical multi-agent RL. First, a strong teacher model converts multimodal GUI trajectories into ordered subtasks, supplying planning annotations that existing GUI trajectory datasets typically lack. Second, these annotations support supervised fine-tuning of two Qwen2.5-VL-7B models: a decomposer that predicts subtasks from the instruction and starting screenshot, and an actor that predicts the next GUI action conditioned on those subtasks. Finally, we jointly train both modules using terminal task success rewards while interacting with web environments, so that the decomposer can adapt its plans to the actor's capabilities while the actor adapts to executing them. We extend Group Relative Policy Optimization (GRPO) through the hierarchical rollout structure: the decomposer proposes $K$ candidate decompositions, and the actor performs $M$ complete environment rollouts for each decomposition. Actor-level advantages compare executions under the same plan, while decomposer-level advantages compare plans by the average terminal reward of their executions, and both policies are updated from this shared batch without dense process rewards.

We evaluate JHiMA-WebRL-14B, our instantiation with a 7B actor and a 7B decomposer, on WebArena-Lite-v2 \citep{liu2025scalecua}, a refined subset of WebArena tasks for purely visual agents. JHiMA-WebRL-14B achieves an average success rate of 37.2\%, compared with 24.0\% for a ScaleCUA-32B agent that we fine-tune with RL on the same training tasks and evaluate under the same protocol. Ablation studies further indicate that jointly updating the decomposer during RL is beneficial relative to updating the actor alone. These results suggest that joint training of role-factorized modules can be a competitive alternative to scaling monolithic agents.

\section{Related Work}

WebArena, VisualWebArena, and WorkArena evaluate agents on web tasks requiring multi-step interaction, visual grounding, state tracking, and long-horizon planning \citep{zhou2024webarena,koh2024visualwebarena,drouin2024workarena}. OSWorld, Windows Agent Arena, and MacArena study the broader setting of desktop and OS-level computer use \citep{xie2024osworld,bonatti2024windowsagentarena,muryn2026macarena}. Strong GUI agents such as UI-TARS and ScaleCUA train largely monolithic policies that map observations and histories directly to actions using large-scale GUI data \citep{qin2025uitars,liu2025scalecua}. These systems demonstrate the importance of data scale and GUI grounding, but require a single policy to learn planning, progress tracking, grounding, and execution jointly, whereas we factorize web interaction into a trainable decomposer and actor.

A broad line of work studies modular or multi-agent LLM systems, including AutoGen, CAMEL, ChatDev, MetaGPT, and MAP \citep{wu2023autogen,li2023camel,qian2024chatdev,hong2024metagpt, webb2025map}. In computer use, Agent S combines manager--worker decomposition with retrieved experience, and Agent S2 updates plans as subtasks are completed \citep{agashe2025agents,agashe2025agents2}. OpAgent coordinates a planner, grounder, reflector, and summarizer, with the reflector's feedback guiding subsequent planning \citep{guo2026opagent}. These mechanisms concern the organization and adaptation of inference-time behavior. Related methods such as ReAct, Reflexion, MiRA, SEAgent, and TreeCUA explore intermediate reasoning, subgoals, feedback, or structured exploration \citep{yao2023react,shinn2023reflexion,wang2026subgoaldriven,sun2025seagent,jiang2026treecua}. JHiMA-WebRL focuses on learning the decomposition interface from annotated trajectories and jointly adapting it with the actor through task rewards.

A complementary line of work studies training across planner--executor interfaces. Plan-and-Act separately fine-tunes planner and executor models using synthetic plans and action trajectories, and adds dynamic replanning at inference time \citep{erdogan2025planandact}. Our approach also uses trajectory-derived planning supervision, then jointly adapts both modules through RL. Navigating WebAI pairs a supervised hierarchical planner with an RL-trained controller \citep{thil2024navigatingwebai}. Planner Matters! optimizes only the planner using trajectory rewards from a VLM judge, freezing the actor and memory manager \citep{wu2026plannermatters}. Its joint-training ablation shares parameters across roles, whereas our decomposer and actor are separate policies. In robotic manipulation, HiRoC aligns an executor with subgoals through supervised fine-tuning, then applies hierarchical GRPO to the executor with a frozen planner, combining task-level and subgoal-level advantages \citep{kong2026hiroc}. SWIRL instead trains both modules through alternating RL phases \citep{lu2025swirl}. In contrast, JHiMA-WebRL computes actor advantages within plans and decomposer advantages across plans, and updates both policies from a shared rollout batch.

PPO and GRPO provide the basic optimization tools for recent LLM and agent RL, while multi-agent RL methods such as MADDPG, COMA, QMIX, and MAPPO study credit assignment across interacting policies \citep{schulman2017ppo,shao2024deepseekmath,lowe2017maddpg,foerster2018coma,rashid2018qmix,yu2022mappo}. For web and GUI agents, WebAgent-R1, WebRL, OpenWebRL, and OpAgent apply online multi-turn RL to browser-based tasks, optimizing a single agent policy from terminal or outcome-based rewards \citep{wei2025webagentr1,qi2024webrl,yang2026openwebrl,guo2026opagent}, while WorkForceAgent-R1 applies rule-based RL to single-step web decisions \citep{zhuang2025workforceagent}. Other works improve credit assignment with process or progress signals, including Web-Shepherd, WebArbiter, TGPO, GraphGPO, and MiRA \citep{chae2025webshepherd,zhang2026webarbiter,chen2025tgpo,cheng2026graphgpo,wang2026subgoaldriven}. We instead extend GRPO's rollout grouping and advantage computation to two jointly trained roles, using only terminal rewards and no dense process rewards.

\section{Method}

\label{sec:method}

We propose JHiMA-WebRL, a joint hierarchical multi-agent reinforcement learning framework for long-horizon web navigation. Our approach factorizes interaction into two specialized but jointly optimized modules: a task decomposer and an actor. The decomposer maps the task instruction and initial screenshot into an ordered sequence of executable subtasks. The actor then conditions on the task, current observation, interaction history, and generated subtasks to produce visually grounded actions in the web interface. Below, we first describe the problem formulation, followed by the three stages of our framework: data generation for training the decomposer, supervised fine-tuning of both the decomposer and actor, and a joint hierarchical multi-agent extension of GRPO.

\subsection{Problem Formulation}
\label{subsec:problem}

We consider an interactive web environment modeled as a partially observed sequential decision problem.
Each task is specified by a natural-language instruction $x$. 
At time step $t$, the agent receives a screenshot $o_t$ of the web interface and
produces an action $a_t \in \mathcal{A}$, such as clicking a page element, typing, scrolling, navigating between webpages, or terminating with an answer.
The environment then transitions to a new state and returns the next observation $o_{t+1}$. A trajectory is denoted by $\tau = (o_0, a_0, o_1, a_1, \ldots, o_T)$, where $T$ is the trajectory length. Each trajectory receives a terminal task reward $R(\tau) \in \{0,1\}$,
where $R(\tau)=1$ if the task is successfully completed and $R(\tau)=0$ otherwise. Success is determined through environment-specific rule-based verification or an LLM judge.

Our modular policy consists of a decomposer policy $\pi_D \equiv \pi_{\theta_D}$ and an actor policy $\pi_A \equiv \pi_{\theta_A}$, parameterized by separate models. 
Given a task instruction $x$ and an initial observation $o_0$, the decomposer samples a decomposition
\begin{equation}
    d = (g_1, g_2, \ldots, g_L) \sim \pi_D(\cdot \mid x, o_0),
\end{equation}
where each $g_\ell$ is a natural-language subtask and $L$ is the number of generated subtasks. 
The actor then generates web-interface actions according to
\begin{equation}
    a_t \sim \pi_A(\cdot \mid x, o_t, h_t, d),
\end{equation}
where $h_t = (a_0,\ldots,a_{t-1})$ denotes the action history.

\subsection{Decomposer Data Generation}
\label{subsec:decomposer_data}

Existing GUI trajectory datasets contain task instructions, screenshots, and action traces, but typically do not provide explicit subtask annotations. 
To train a task decomposer, we construct a new decomposition dataset from trajectory annotations in ScaleCUA \citep{liu2025scalecua}. 
ScaleCUA provides large-scale multimodal GUI data across platforms, including grounding annotations and human or agent trajectories. 

For each trajectory, we provide a strong teacher model, GPT-5.2-Codex \citep{openai2025gpt52codex}, with the original task instruction $x$ and starting screenshot $o_0$, together with the full sequence of user or agent actions $(a_0,\ldots,a_{T-1})$ and corresponding screenshots $(o_0,\ldots,o_T)$. A subgoal-generation guideline prompt, included in Appendix~\ref{app:subtask_prompt}, guides the annotation process.

The teacher produces an ordered list of executable subtasks $d=(g_1,\ldots,g_L)$ and optional descriptions of unnecessary or redundant actions in the trajectory. This process yields approximately $76$K decomposer training examples; dataset statistics are provided in Table~\ref{tab:dataset-stats}, and qualitative examples are shown in Appendix~\ref{app:qualitative}.
The resulting automated pipeline converts low-level multimodal GUI demonstrations into high-level planning supervision, enabling us to train an open-source decomposer rather than relying on a proprietary teacher at inference time.


\subsection{Supervised Fine-Tuning}
\label{subsec:sft}

We first train both modules with supervised fine-tuning before reinforcement learning.

\paragraph{Decomposer SFT.}
The decomposer is initialized from the Qwen2.5-VL-7B model and trained on the generated subtask dataset.
Given the task instruction and initial screenshot, the model is trained to predict the teacher-generated decomposition:
\begin{equation}
    \mathcal{L}_{D}^{\mathrm{SFT}}
    =
    - \sum_{n=1}^{N_d}
    \log
    \pi_{\theta_D}
    \left(
        d^{(n)}
        \mid
        x^{(n)}, o_0^{(n)}
    \right),
\end{equation}
where $N_d$ is the number of decomposer examples.

\paragraph{Actor SFT.}
The actor is also initialized from Qwen2.5-VL-7B and fine-tuned using supervised learning on released ScaleCUA trajectories augmented with the teacher-generated subtasks described in Section~\ref{subsec:decomposer_data}. For each trajectory, the actor receives the generated subtask list together with the task, current screenshot, and previous actions.
It is then trained to predict the demonstrated next action:
\begin{equation}
    \mathcal{L}_{A}^{\mathrm{SFT}}
    =
    - \sum_{n=1}^{N_A}
    \sum_{t=0}^{T_n-1}
    \log
    \pi_{\theta_A}
    \left(
        a_t^{(n)}
        \mid
        x^{(n)}, o_t^{(n)}, h_t^{(n)}, d^{(n)}
    \right),
\end{equation}
where $N_A$ is the number of actor trajectories. 
This trains the actor to use decompositions as an explicit planning context while learning GUI grounding and action generation from the released ScaleCUA trajectories.

\subsection{Joint Hierarchical Multi-Agent GRPO}
\label{sec:jhima_grpo}

After supervised fine-tuning, the joint RL stage of JHiMA-WebRL optimizes the decomposer and actor using GRPO \citep{shao2024deepseekmath}, extended to their hierarchical interaction. We retain the GRPO clipped policy objectives and compute role-specific advantages from hierarchical rollout groups to assign credit to planning and execution.
Standard web-agent RL methods often optimize a single policy from terminal trajectory rewards \citep{wei2025webagentr1,qi2024webrl,yang2026openwebrl}. 
In contrast, our setting contains two interacting policies with different roles and action spaces: the decomposer generates a high-level plan, and the actor executes low-level web-interface actions conditioned on that plan.

For a task $x$ with initial observation $o_0$, the decomposer first samples
$K$ candidate decompositions $d_i \sim \pi_{\theta_D}(\cdot \mid x,o_0)$, $i=1,\ldots,K$.
For each $d_i$, the actor then generates $M$ independent multi-turn
trajectories $\tau_{i,j} \sim \pi_{\theta_A}(\cdot \mid x,o_0,d_i)$, $j=1,\ldots,M$, by interacting with the environment.
We denote the actions of $\tau_{i,j}$ by $\left(a_{i,j,0},\ldots,a_{i,j,T_{i,j}-1} \right)$, where $T_{i,j}$ is the number of steps and $a_{i,j,k}$ is the action generated at environment interaction step
$k$. Because each action is produced autoregressively by the VLM, we further
write $a_{i,j,k}=\left(a_{i,j,k,1},\ldots,a_{i,j,k,|a_{i,j,k}|}
    \right)$,
where $u$ indexes the generated tokens $a_{i,j,k,u}$ within the action.
Each actor trajectory receives a terminal binary reward $R_{i,j} = R(\tau_{i,j}) \in \{0,1\}$.

\paragraph{Actor-level advantage.}
For a fixed decomposition $d_i$, we compute the group-relative actor advantage
by comparing the $M$ downstream trajectories generated under the same
decomposition:
\begin{equation}
    A^{A}_{i,j}
    =
    \frac{
        R_{i,j}
        -
        \frac{1}{M}\sum_{m=1}^{M}R_{i,m}
    }{
        \operatorname{std}_{m}
        \left[R_{i,m}\right]
        + \delta
    },
    \label{eq:actor_advantage}
\end{equation}
where $\delta>0$ is a small constant for numerical stability.
Thus, all action tokens in trajectory $\tau_{i,j}$ share the same
trajectory-level advantage $A^{A}_{i,j}$.

Following the M-GRPO objective
\citep{wei2025webagentr1}, which applies the loss per token, we define the actor importance-sampling ratio for
token $u$ of action $a_{i,j,k}$ as
\begin{equation}
    r^{A}_{i,j,k,u}(\theta_A)
    =
    \frac{
        \pi_{\theta_A}
        \left(
            a_{i,j,k,u}
            \mid
            c^{A}_{i,j,k},
            a_{i,j,k,<u}
        \right)
    }{
        \pi_{\theta_A^{\mathrm{old}}}
        \left(
            a_{i,j,k,u}
            \mid
            c^{A}_{i,j,k},
            a_{i,j,k,<u}
        \right)
    },
    \label{eq:actor_ratio}
\end{equation}
where $c^{A}_{i,j,k}=(x, o_{i,j,k}, h_{i,j,k}, d_i)$ contains the task instruction, current observation, action history available at environment step
$k$, and decomposition $d_i$.

The clipped token-level actor objective is
\begin{align}
\mathcal{L}_{A}
=
-\frac{1}{KM}
\sum_{i=1}^{K}
\sum_{j=1}^{M}
\frac{1}{T_{i,j}}
\sum_{k=0}^{T_{i,j}-1}
\frac{1}{|a_{i,j,k}|}
\sum_{u=1}^{|a_{i,j,k}|}
\Big[
&
\min\big(
r^{A}_{i,j,k,u}(\theta_A) A^{A}_{i,j},
\nonumber\\[-1mm]
&
\operatorname{clip}
(r^{A}_{i,j,k,u}(\theta_A),
1-\epsilon_A,
1+\epsilon_A)
A^{A}_{i,j}
\big)
\nonumber\\
&
-\beta_A
D_{\mathrm{KL}}^{A}(i,j,k,u)
\Big],
\label{eq:actor_grpo}
\end{align}
where $\epsilon_A$ is the clipping parameter and $D_{\mathrm{KL}}^{A}(i,j,k,u)$ is the per-token KL divergence between $\pi_{\theta_A}$ and the reference (SFT) actor, weighted by $\beta_A$.

\paragraph{Decomposer-level advantage.}
To evaluate decomposition quality, we first aggregate the downstream rewards of
the $M$ actor trajectories conditioned on $d_i$:
\begin{equation}
    S_i
    =
    \frac{1}{M}
    \sum_{j=1}^{M}
    R_{i,j}.
    \label{eq:decomposition_score}
\end{equation}
The decomposer advantage is then computed relative to the $K$ candidate
decompositions sampled for the same task:
\begin{equation}
    A^{D}_{i}
    =
    \frac{
        S_i -
        \frac{1}{K}\sum_{i'=1}^{K}S_{i'}
    }{
        \operatorname{std}_{i'}[S_{i'}]
        +\delta
    }.
    \label{eq:decomposer_advantage}
\end{equation}

Because $d_i$ is itself an autoregressively generated sequence, we define the token-level ratio $r^{D}_{i,u}(\theta_D)=\pi_{\theta_D}(d_{i,u}\mid x,o_0,d_{i,<u})/\pi_{\theta_D^{\mathrm{old}}}(d_{i,u}\mid x,o_0,d_{i,<u})$ for its $u$-th token, analogously to Eq.~\eqref{eq:actor_ratio}. The token-level decomposer objective is
\begin{align}
\mathcal{L}_{D}
=
-\frac{1}{K}
\sum_{i=1}^{K}
\frac{1}{|d_i|}
\sum_{u=1}^{|d_i|}
\Big[
&
\min\big(
r^{D}_{i,u}(\theta_D) A^{D}_{i},
\nonumber\\[-1mm]
&
\operatorname{clip}
(r^{D}_{i,u}(\theta_D),
1-\epsilon_D,
1+\epsilon_D)
A^{D}_{i}
\big)
\nonumber\\
&
-\beta_D
D_{\mathrm{KL}}^{D}(i,u)
\Big],
\label{eq:decomposer_grpo}
\end{align}
with $\epsilon_D$, $\beta_D$, and $D_{\mathrm{KL}}^{D}$ defined analogously to those for the actor.

Thus, the downstream execution quality of a decomposition determines a single
group-relative advantage $A_i^D$, which is propagated to all generated tokens
of that decomposition. This enables the decomposer to learn which planning
outputs are most useful to the actor while retaining token-level policy
optimization.

Finally, the two modules are optimized jointly according to
\begin{equation}
    \mathcal{L}_{\mathrm{JHiMA}}
    =
    \mathcal{L}_{A}
    +
    \mathcal{L}_{D}.
    \label{eq:joint_objective}
\end{equation}
Since $\theta_A$ and $\theta_D$ are disjoint, this sum amounts to updating each module with its own objective, using the same rollout batch.
Equations~\eqref{eq:actor_advantage}--\eqref{eq:decomposer_grpo} give the standard GRPO form; in our experiments we remove the standard-deviation and response-length normalization following \citet{liu2025understandingr1zero} (Section~\ref{sec:experimental_details}).

Appendix~\ref{app:interpretation} shows that, without standard-deviation normalization, $A^D_i$ yields an unbiased (up to a positive scale) baselined policy-gradient estimate for the decomposer under the shared task reward.

\section{Experiments}

\subsection{Datasets}
\label{sec:datasets}

The task decomposer was trained with supervised fine-tuning using approximately $76$K task-decomposition examples, which were generated from ScaleCUA trajectories using GPT-5.2-Codex as described in Section~\ref{subsec:decomposer_data}. The actor was fine-tuned on all ScaleCUA trajectories augmented with the teacher-annotated subtasks, along with grounding data spanning web, mobile, and desktop platforms, resulting in 2,535,552 training examples. Unlike standard ScaleCUA SFT, the actor thus receives a task decomposition in its context as a planning scaffold.

Both modules were then jointly trained through online reinforcement learning using the training tasks associated with WebArena-Lite, following the setup of WebAgent-R1 \citep{wei2025webagentr1}. WebArena contains $812$ interactive web tasks spanning five domains---Reddit, GitLab, e-commerce administration, maps, and online shopping. WebArena-Lite \citep{liu2025visualagentbench} selects a cleaned subset of $165$ tasks for evaluation, leaving $647$ tasks from the original WebArena benchmark for reinforcement-learning interaction. No WebArena-Lite evaluation task (and hence no WebArena-Lite-v2 task) is used during SFT or RL. We also employed the flexible evaluation protocol of \citet{liu2025scalecua}, including LLM-based fuzzy matching using GPT-5.2-Codex, to compute the binary task outcome reward at the end of web interaction.

For evaluation, we used WebArena-Lite-v2, introduced by ScaleCUA \citep{liu2025scalecua}. WebArena-Lite-v2 is a refinement of the evaluation subset of WebArena-Lite designed specifically for evaluating purely visual GUI agents; it removes $11$ tasks that require multi-tab interaction and revises task instructions and evaluators. It contains $154$ tasks across five website environments: Shopping, ShoppingAdmin, Reddit, GitLab, and OpenStreetMap. The benchmark includes heterogeneous task objectives such as question answering, navigation, page-content matching, and state-changing operations, with success determined by the flexible evaluation criteria of \citet{liu2025scalecua}.

\subsection{Baselines}
\label{sec:baselines} 

We compared against native agents at several model scales ranging from 7B to 72B, allowing comparisons between our 14B modular system and both smaller and larger monolithic systems. We also compared against multi-agent systems consisting of a proprietary planner (GPT-4o) with a grounding module. These provide comparisons with systems that separate planning from visual grounding, but do not jointly train a learned decomposer and actor as in JHiMA-WebRL-14B. The UI-TARS and GPT-4o planner--grounder results are taken from ScaleCUA \citep{liu2025scalecua} (15-step budget). The ScaleCUA-7B/32B SFT and GUI-Libra-8B results are taken from the re-evaluation of \citet{yang2026guilibra}, which uses a 15-step budget, averages four runs, and replaces the GPT-4o fuzzy-match judge with GPT-5. Under its original evaluation setup, ScaleCUA reports higher scores for its own models (37.7\% for ScaleCUA-7B and 44.2\% for ScaleCUA-32B at 15 steps). We also trained a ScaleCUA-32B RL baseline, specifically fine-tuning the SFT model on the WebArena training set, using standard multi-turn GRPO with 7 rollouts and a maximum of 15 interaction steps. For a fair comparison with our method, we only included baselines that do not use HTML or DOM information.

\subsection{Experimental Details}
\label{sec:experimental_details} 

We trained the 7B decomposer and 7B actor of JHiMA-WebRL-14B in two stages: SFT, followed by joint hierarchical RL. The decomposer was initialized from Qwen2.5-VL-7B and fine-tuned on the approximately $76$K decomposition examples generated from ScaleCUA multimodal trajectories, on 8 NVIDIA H100 GPUs with 80 GB of memory each via
full-parameter fine-tuning. We trained for \texttt{1} epoch with a learning rate of \texttt{1e-5} and a \texttt{cosine} learning-rate schedule with a warmup ratio of \texttt{0.03}. The per-device batch size was \texttt{2}, with gradient accumulation over \texttt{4} steps, resulting in an effective global batch size of \texttt{64}. We used DeepSpeed \citep{rajbhandari2020zero} for distributed training with ZeRO-3,
along with gradient checkpointing, bf16 mixed-precision training, gradient clipping to a maximum norm of \texttt{1}, and a maximum token length of \texttt{30720}. The actor was also initialized from Qwen2.5-VL-7B and fine-tuned with the same hyperparameters as above, using released ScaleCUA trajectories augmented with generated subtasks.

After SFT, both the actor and decomposer were jointly optimized through LoRA fine-tuning using GRPO, with standard-deviation and response-length normalization removed to reduce optimization bias, following \citet{liu2025understandingr1zero}. We used LoRA rank $r=\texttt{32}$ and scaling factor $\alpha=\texttt{64}$ for both modules, with a separate adapter for each. For each task, the decomposer sampled $K=\texttt{3}$ candidate decompositions. For every decomposition, the actor generated $M=\texttt{7}$ complete trajectories. For both, we used \texttt{0.8} as the sampling temperature, and we allowed a maximum of \texttt{15} interaction steps per trajectory. We optimized the LoRA parameters with a learning rate of \texttt{1e-5}. The GRPO updates used clipping parameters $\epsilon_A=\epsilon_D=\texttt{0.2}$, KL coefficients $\beta_A=\beta_D=\texttt{0}$ (i.e., no KL penalty), and \texttt{2} optimization steps per rollout batch. We used \texttt{32} gradient accumulation steps and trained for \texttt{1} epoch over the $647$ training tasks. We used the efficient parallel trajectory rollout generation of M-GRPO \citep{wei2025webagentr1} for each task decomposition, with training taking approximately 720 GPU-hours on a cluster of 8 H100 GPUs with 80 GB of memory each.

\paragraph{Evaluation.} We report the task success rate on all $154$ WebArena-Lite-v2 tasks, averaged over $4$ independent evaluation runs. We sampled one decomposition at test time using a temperature of 0.0 and allowed a maximum of 15 interaction steps, consistent with the baselines.



\begin{table}[h]
\caption{Success rates (\%) on WebArena-Lite-v2. Our results are averaged over 4 runs. $^{*}$Reported by ScaleCUA \citep{liu2025scalecua}; $^{\dagger}$evaluated by GUI-Libra \citep{yang2026guilibra}; $^{\ddagger}$trained and evaluated by us. All rows use a 15-step budget, but the fuzzy-match judge differs across sources (Section~\ref{sec:baselines}). Average is computed over all 154 tasks (task-weighted), not as the mean of the domain scores. Dashes indicate domain scores not reported.}
\label{tab:main_results}
\centering
\resizebox{\linewidth}{!}{
\begin{tabular}{lcccccc}
\textbf{Method} & \textbf{GitLab} & \textbf{Map} & \textbf{Reddit} & \textbf{Shopping} & \textbf{ShoppingAdmin} & \textbf{Average} \\
\hline
UI-TARS-1.5-7B$^{*}$
    & -- & -- & -- & -- & -- & 20.8 \\

UI-TARS-72B-DPO$^{*}$
    & -- & -- & -- & -- & -- & 23.4 \\
ScaleCUA-7B SFT$^{\dagger}$
    & 28.3 & 15.4 & 27.6 & 18.8 & 30.7 & 23.9 \\

ScaleCUA-32B SFT$^{\dagger}$
    & 34.2 & 10.6 & 26.3 & 16.5 & 33.6 & 24.0 \\

ScaleCUA-32B RL$^{\ddagger}$
    & 35.0 & 10.6 & 21.0 & 24.4 & 25.7 & 24.0 \\

GUI-Libra-8B$^{\dagger}$
    & 31.7 & \textbf{17.3} & 35.5 & 26.1 & 25.0 & 26.6 \\

GPT-4o Planner + UI-TARS-1.5-7B Grounder$^{*}$
    & -- & -- & -- & -- & -- & 22.6 \\
GPT-4o Planner + ScaleCUA-7B Grounder$^{*}$
    & -- & -- & -- & -- & -- & 28.6 \\
\textbf{JHiMA-WebRL-14B (Ours)}
    & \textbf{47.5} & 12.5 & \textbf{60.5} & \textbf{29.0} & \textbf{44.3} & \textbf{37.2} \\
\hline
\end{tabular}
}
\end{table}

\subsection{Results}

As shown in Table~\ref{tab:main_results}, JHiMA-WebRL-14B achieved the highest average success rate among the listed methods, 37.2\% over four runs, and the highest scores on GitLab (47.5\%), Reddit (60.5\%), Shopping (29.0\%), and ShoppingAdmin (44.3\%). The most controlled comparison is with ScaleCUA-32B RL, which we trained on the same RL tasks with the same step budget and evaluated under the same protocol: JHiMA-WebRL-14B outperformed it by 13.2 percentage points with fewer total parameters. Among results from prior work, JHiMA-WebRL-14B exceeded the GPT-4o + ScaleCUA-7B planner--grounder system (28.6\%), GUI-Libra-8B (26.6\%), and UI-TARS-72B-DPO (23.4\%); however, these numbers come from evaluation setups that differ from ours, and ScaleCUA reports 44.2\% for ScaleCUA-32B under its original setup (Section~\ref{sec:baselines}). The Map domain is the exception: JHiMA-WebRL-14B (12.5\%) trails GUI-Libra-8B (17.3\%) and ScaleCUA-7B SFT (15.4\%). Figure~\ref{fig:map_failure} shows an example Map failure in which the agent repeatedly toggles the zoom controls until the 15-step budget is exhausted. Within our controlled comparisons, these results suggest that learned role factorization with smaller, jointly trained models is a competitive alternative to scaling a monolithic agent.

\begin{table}[h]
\caption{Ablation success rates (\%) on WebArena-Lite-v2, averaged over 4 runs (task-weighted average). ``Rollouts'' denotes the number of complete trajectories per training task.}
\label{tab:ablation_decomposer}
\centering
\resizebox{\linewidth}{!}{
\begin{tabular}{lcccccc}
\textbf{Method} & \textbf{GitLab} & \textbf{Map} & \textbf{Reddit} & \textbf{Shopping} & \textbf{ShoppingAdmin} & \textbf{Average} \\
\hline
\textbf{JHiMA-WebRL-14B (Ours)}
     & \textbf{47.5} & 12.5 & \textbf{60.5} & 29.0 & 44.3 & \textbf{37.2} \\
\quad w/o decomposer RL
    & 37.5 &10.6 & 54.0 & 25.6 & 45.7 & 33.4 \\

\quad w/o decomposer (actor-only RL, 21 rollouts)
    & 44.2 & 14.4 & 51.3 & 26.7 & 43.6 & 34.9 \\

\quad w/o decomposer (actor-only RL, 7 rollouts)
    & 39.2 & \textbf{15.4} & 48.7 & \textbf{30.1} & 45.7 & 35.2 \\
\quad w/o RL (SFT decomposer + SFT actor)
    & 43.3 & 8.6 & 59.2 & 25.6 & \textbf{46.4}  & 35.1  \\
\hline
\end{tabular}
}
\end{table}

\subsection{Ablation Study}

We analyzed the contribution of the components of JHiMA-WebRL-14B through ablations (Table~\ref{tab:ablation_decomposer}). First, we trained only the actor with RL while keeping the SFT decomposer frozen (`w/o decomposer RL'), using the same number of rollouts per task ($K{=}3$, $M{=}7$) as JHiMA-WebRL-14B. This reduced the average success rate to 33.4\%, a drop of 3.8 percentage points, indicating that updating the decomposer jointly with the actor matters. Second, to assess the role of the decomposer, we trained the actor alone using standard multi-turn GRPO without subtask conditioning. The actor was initialized from Qwen2.5-VL-7B and fine-tuned using supervised learning on released ScaleCUA trajectories and grounding data, without any subtasks. We trained two variants: one with the same total number of rollouts per task as JHiMA-WebRL-14B (21), and one with 7 rollouts, matching the per-decomposition group size of our actor. The two variants reached 34.9\% and 35.2\%, i.e., 2.3 and 2.0 percentage points below JHiMA-WebRL-14B. Finally, we removed RL entirely and evaluated the SFT decomposer and SFT actor together (`w/o RL'), which reduced the average success rate by 2.1 percentage points. Notably, actor-only RL with a frozen SFT decomposer (33.4\%) scored below this SFT-only system (35.1\%), whereas joint RL improved on it; the SFT-only modular system also performed on par with the decomposer-free actors trained with RL (34.9\% and 35.2\%).



\begin{figure}[h]
    \centering

    \begin{subfigure}[t]{0.33\linewidth}
        \centering
        \includegraphics[width=\linewidth]{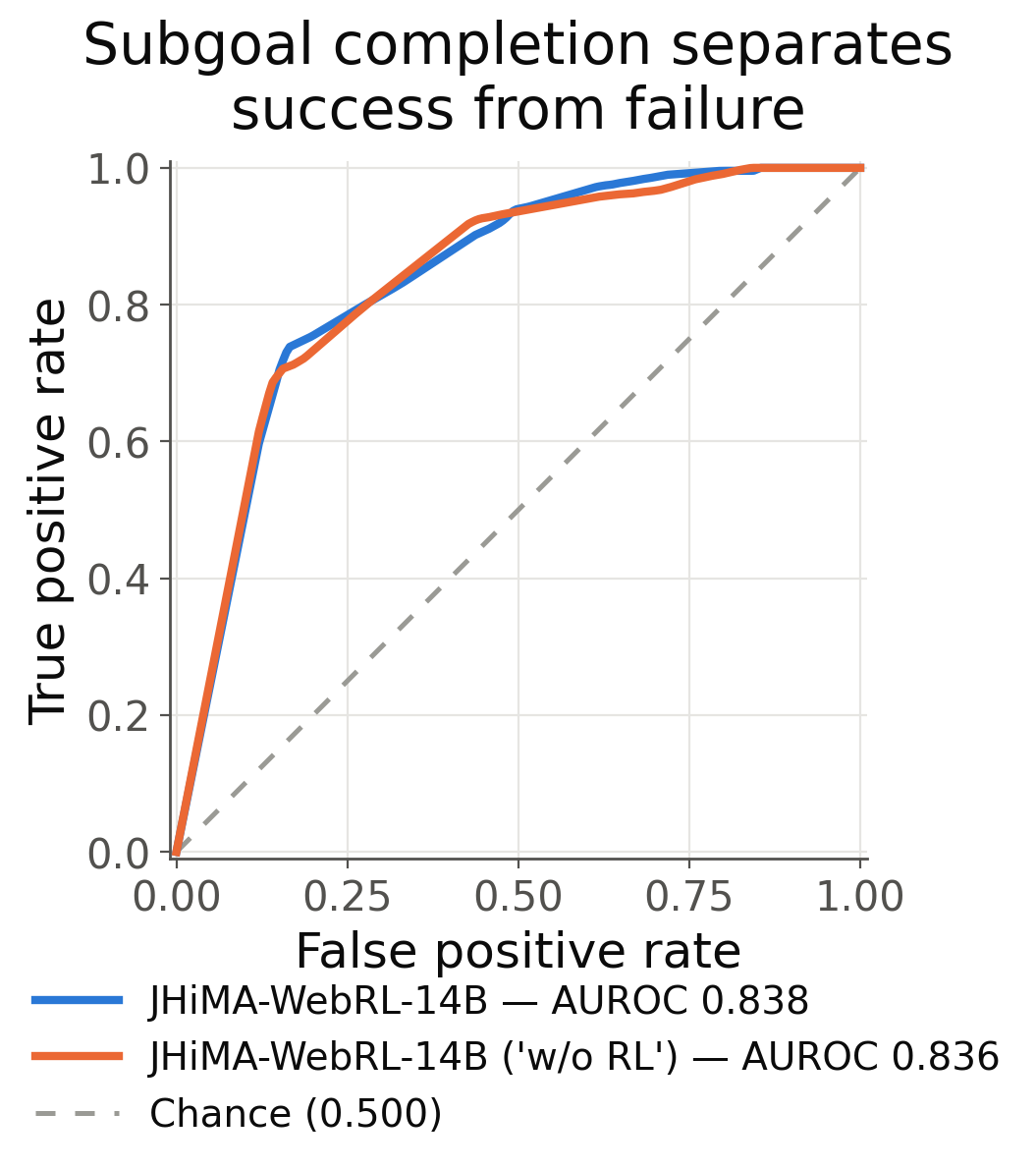}
        \caption{
        ROC curve using the fraction of completed subgoals as a score for
        predicting final task success.
        }
        \label{fig:subgoal_quality_auc}
    \end{subfigure}
    \hfill
    \begin{subfigure}[t]{0.5\linewidth}
        \centering
        \includegraphics[width=\linewidth]{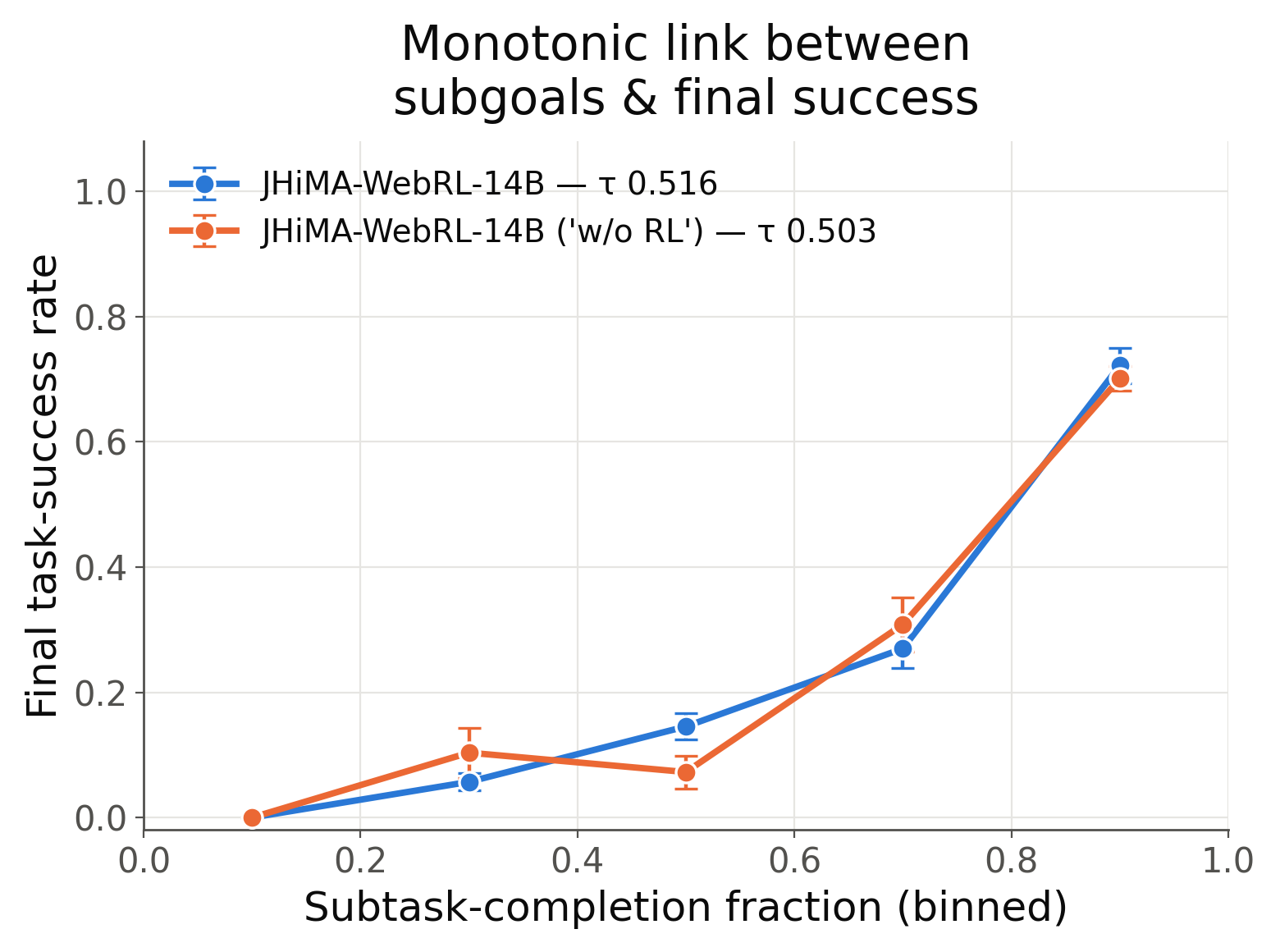}
        \caption{
        Empirical task success rate as a function of subgoal completion.
        }
        \label{fig:subgoal_quality_monotonicity}
    \end{subfigure}

    \caption{
    \textbf{Subgoal quality analysis.}
    (a) Discriminative power of subgoal completion for predicting terminal
    task success. (b) Empirical task success versus binned subgoal-completion fraction. Both panels compare JHiMA-WebRL-14B with the SFT-only system (`w/o RL').
    }
    \label{fig:subgoal_quality}
\end{figure}

\subsection{Subgoal Quality Analysis}
\label{sec:subgoal_quality}

We also analyzed whether the generated decompositions provide meaningful intermediate structure. Following the subgoal-quality analysis of
\citet{wang2026subgoaldriven}, we measured whether partial subgoal completion was
predictive of eventual task success and whether task success increased
monotonically as more subgoals were completed.

For each WebArena-Lite-v2 trajectory from the four evaluation runs, we used GPT-5.2-Codex to judge whether
each subgoal generated by the decomposer was completed, based on the sequence of predicted actions and resulting screenshots, and computed the fraction of
generated subgoals that were completed.
We first evaluated whether the fraction of completed subgoals discriminated successful
trajectories from failed ones. We treated the fraction as a continuous
prediction score for the binary task outcome and computed the area under the
receiver operating characteristic curve (AUROC).
As shown in Figure~\ref{fig:subgoal_quality_auc}, the subgoals generated by
JHiMA-WebRL-14B achieved an AUROC of
0.838, well above chance (0.5) and comparable to the SFT-only system (`w/o RL', 0.836). Progress through the generated decompositions is therefore predictive of downstream task success, both before and after joint RL.

A useful decomposition should also provide a monotone notion of progress, so we grouped
trajectories by subgoal completion fraction and computed the
empirical success rate per group. For JHiMA-WebRL-14B, empirical task success increases monotonically with subgoal
completion fraction (Figure~\ref{fig:subgoal_quality_monotonicity}), with Kendall's rank correlation
coefficient $\tau=0.516$, compared with $\tau=0.503$ without RL. Together with the AUROC analysis, these
results suggest that the generated subgoals act as intermediate milestones associated with downstream task completion.

\section{Limitations}
\label{sec:limitations}

Our study has several limitations. (i) \emph{Train--test overlap:} RL is performed on the remaining WebArena tasks, which are disjoint from the evaluation tasks but can share intent templates and websites with them; results thus reflect in-environment generalization, not transfer to unseen websites. (ii) \emph{Supervision and cost:} decomposition supervision is generated by a proprietary teacher (GPT-5.2-Codex) without human verification of annotation quality, and hierarchical rollouts require $K{\times}M$ complete episodes per task, which is costly in live environments. (iii) \emph{Open-loop planning:} the decomposer plans once from the initial observation and cannot revise the plan during execution. (iv) \emph{Coarse credit:} terminal rewards assign the same advantage to every token of a decomposition or trajectory, without identifying which subtask or action caused success or failure. (v) \emph{Evaluation scope:} we evaluate on a single benchmark of 154 tasks, and the ablation differences (2.0--3.8 percentage points) are of a size where run-to-run variance matters.

\section{Conclusion and Future Directions}

We introduced JHiMA-WebRL, a framework for jointly training a task decomposer and a visually grounded actor. The framework generates subtask supervision from multimodal GUI trajectories to train the decomposer and extends GRPO with hierarchical rollout grouping and role-specific advantages. JHiMA-WebRL-14B achieved an average success rate of 37.2\% on WebArena-Lite-v2, 13.2 percentage points above a ScaleCUA-32B RL baseline trained and evaluated under the same protocol. In ablations, removing decomposer updates during RL, removing the decomposer together with subtask conditioning, or removing RL each lowered the average success rate, by 2.0--3.8 percentage points.

Our work suggests several future directions. First, the decomposer could be made \emph{adaptive}, revising subtasks as execution deviates from the plan. Second, joint optimization can be combined with intermediate progress signals such as milestone rewards \citep{wang2026subgoaldriven} or graph-based credit assignment \citep{cheng2026graphgpo}. Finally, jointly trained decomposer--actor systems could be extended to cross-domain and live-web settings, where online RL has recently been shown to be feasible \citep{yang2026openwebrl}. This would test whether modular role factorization transfers better than monolithic policies.

\subsection*{AI use statement}

Generative AI is part of our method: GPT-5.2-Codex is used as a teacher model to generate subtask annotations from ScaleCUA trajectories (Section~\ref{subsec:decomposer_data}) and as a judge of subgoal completion in the analysis of Section~\ref{sec:subgoal_quality}; the evaluation protocol of \citet{liu2025scalecua} also uses an LLM for fuzzy answer matching. We used Claude Opus 4.8 for coding and GPT-5.6 and GPT-6 to assist with drafting the paper and reviewing the literature. We have reviewed all AI-assisted work and take responsibility for the final content of this work, including text, claims, and artifacts produced with the aid of generative AI.



\subsection*{Reproducibility statement}

Section~\ref{sec:method} specifies the full training objective, including the hierarchical advantages (Eqs.~\ref{eq:actor_advantage}--\ref{eq:decomposer_advantage}) and the clipped objectives for both modules. Section~\ref{sec:datasets} describes the training and evaluation data, and Section~\ref{sec:experimental_details} lists all SFT and RL hyperparameters. The exact prompt used to generate decomposition supervision is given in Appendix~\ref{app:subtask_prompt}. We will also release the code, the decomposition dataset, and model checkpoints.



\bibliography{iclr2027_conference}
\bibliographystyle{iclr2027_conference}

\newpage
\appendix
\section*{Appendix}

\begin{table}[h]
\centering
\caption{Subtask-decomposition dataset statistics.}
\label{tab:dataset-stats}
\begin{tabular}{lrrr}
\toprule
Platform & \# Examples & Avg. \# Subtasks & Frac. w/ Unnecessary Actions \\
\midrule
Windows &  5{,}873 & 2.68 & 1.2\% \\
macOS   &  3{,}845 & 2.43 & 1.1\% \\
Linux   & 25{,}122 & 2.20 & 1.8\% \\
Android & 17{,}245 & 2.84 & 1.6\% \\
Web     & 24{,}311 & 2.73 & 1.5\% \\
\midrule
Total   & 76{,}396 & 2.56 & 1.6\% \\
\bottomrule
\end{tabular}
\end{table}

\section{Subtask Generation Prompt}
\label{app:subtask_prompt}

The following prompt is provided together with the task instruction and
interleaved trajectory screenshots. Images are inserted as multimodal image
content immediately after the corresponding textual markers.


\begin{lstlisting}[style=promptstyle]
You are an AI assistant tasked with generating high level subtasks that highlight key subgoals necessary to complete a task.

Task: {TASK}

Starting Image URL:
{STARTING_IMAGE}

Below are the actions taken by the user at each step to perform the task

Step 1: {STEP_1}
Image URL after step 1:
{IMAGE_AFTER_STEP_1}

Step 2: {STEP_2}
Image URL after step 2:
{IMAGE_AFTER_STEP_2}

...

Step N: {STEP_N}
Image URL after step N:
{IMAGE_AFTER_STEP_N}

Guidelines for Subtask Generation
    1. Identify Essential High-Level Subgoals:
    - A subgoal should represent a significant step involving user action that leads to
    noticeable page transitions or meaningful changes in system state (i.e Image).
    - Consolidate closely related user actions (such as applying multiple filters or selecting several options) into a single subgoal, rather than separate subtasks for each action.
    - Prioritize only the most critical actions necessary for meaningful progression,
    avoiding the inclusion of minor or unnecessary steps (e.g., scroll, hover).
    
    2. Provide a Concise Subgoal Analysis:
    - Before creating the subtasks, offer a brief paragraph summarizing the main subgoals,
    emphasizing significant transitions or page-level actions.
    
    3. Ensure Clear Goal:
    - If multiple related actions occur (e.g., setting filters 1, 2, and 3), combine them into one
    subgoal with clear criteria verifying all required conditions.
    - The subtasks should contain only essential steps, explicitly excluding unnecessary
    actions, and should not exceed five critical subtasks. It is not necessary to use all five
    subtasks if fewer steps adequately represent the essential subgoals.
    
    Finally, based on the generated subtasks and subgoal analysis, report if there are any unnecessary or redundant actions which aren't required to perform the task.
    If all the actions are important say <NONE>, else say <YES> alongwith the unnecessary actions.
    
    ### Output Format
    Before generating the subtasks, first produce a concise subgoal analysis in a single
    paragraph summarizing the required actions. Then, based on this, generate the subtasks
    following the format below:
    
    [SUBGOAL ANALYSIS]
    [One-paragraph summary explaining the key subgoals and their logical
    sequence in task completion.]
    
    [SUBTASKS]
    SUBTASK X:
    [Short title of the action/goal]
    - Goal:
    [Brief description of the subgoal at this stage, emphasizing the
    purpose of the action.]
    
    [UNNECESSARY ACTIONS]
    [NONE or YES alongwith the unnecessary actions]
\end{lstlisting}

\section{Policy-Gradient Interpretation of the Hierarchical Advantages}
\label{app:interpretation}

Both modules share the single objective
$J(\theta_D,\theta_A)=\mathbb{E}_{d\sim\pi_{\theta_D}}\,\mathbb{E}_{\tau\sim\pi_{\theta_A}(\cdot\mid x,o_0,d)}[R(\tau)]$,
so the setting is fully cooperative with a shared reward.
By the policy-gradient theorem, $\nabla_{\theta_D}J=\mathbb{E}_{d}\big[\nabla_{\theta_D}\log\pi_{\theta_D}(d\mid x,o_0)\,(Q(d)-b)\big]$ with $Q(d)=\mathbb{E}_{\tau}[R(\tau)\mid d]$ and any baseline $b$ that does not depend on $d$.
The score $S_i$ in Eq.~\eqref{eq:decomposition_score} is an unbiased Monte Carlo estimate of $Q(d_i)$ under the current actor; the $M$ inner rollouts reduce its variance relative to scoring a decomposition by a single rollout.
Because the group mean $\bar S$ includes $S_i$, we have $S_i-\bar S=\frac{K-1}{K}\,(S_i-\bar S_{-i})$, where the leave-one-out mean $\bar S_{-i}$ depends only on the other $K-1$ independently sampled decompositions and their rollouts, and is therefore a valid baseline for $d_i$.
Hence, without standard-deviation normalization, the decomposer advantage gives an unbiased policy-gradient estimate up to the positive factor $(K-1)/K$; dividing by the group standard deviation, as in standard GRPO, rescales the estimate per task and no longer preserves this property exactly.
Likewise, conditioning the actor baseline on $d_i$ (Eq.~\eqref{eq:actor_advantage}) removes the reward variance attributable to plan quality from the actor's advantage, so that actor updates compare executions of the same plan.
Because both policies change simultaneously, each module's learning problem is non-stationary; we rely on the PPO-style clipping and a small number of optimization steps per batch to keep per-iteration policy changes small.

\section{Qualitative Examples}
\label{app:qualitative}

\begin{figure}[h]
    \centering
    \includegraphics[width=\linewidth]{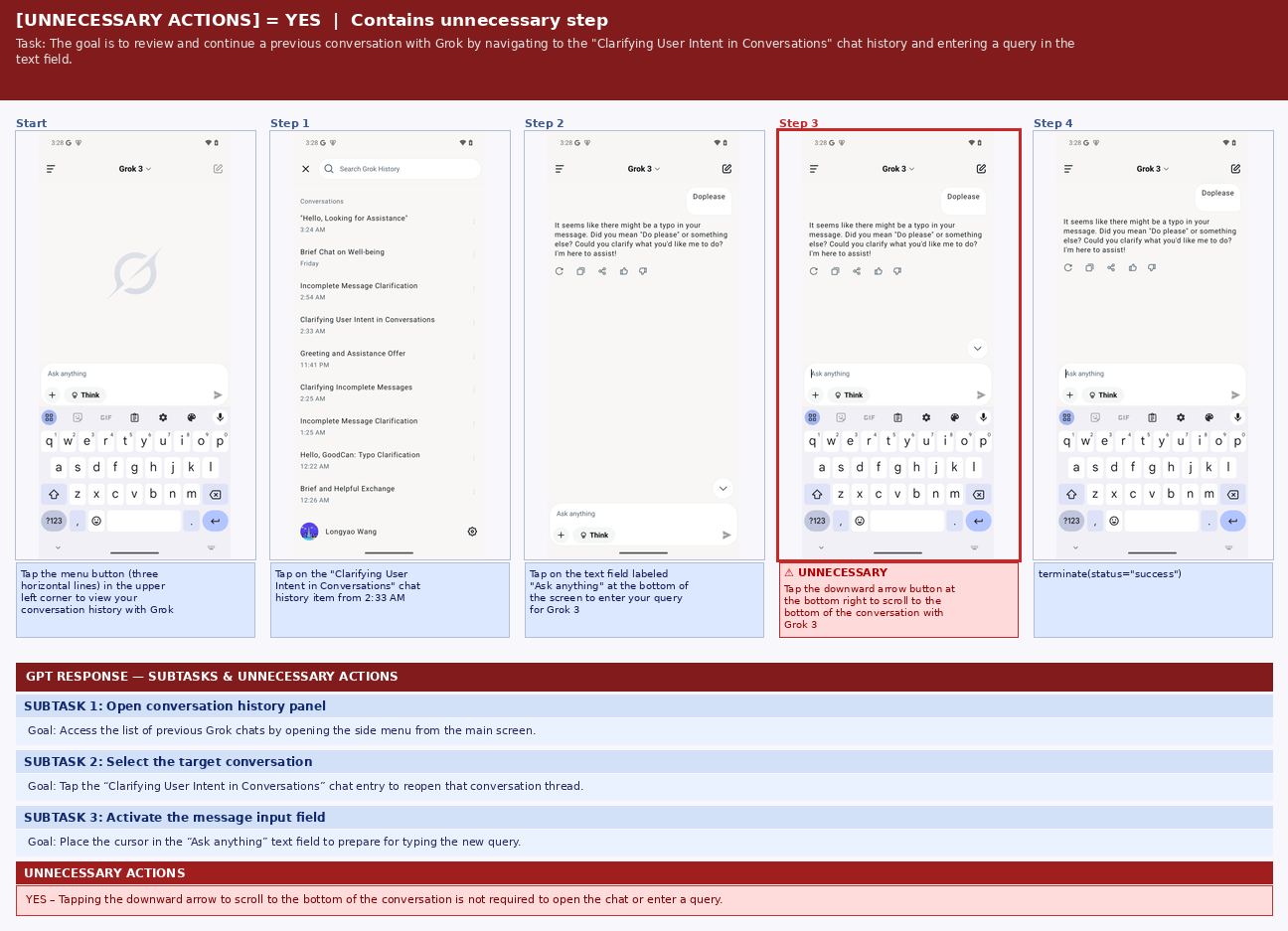}
    \caption{Teacher annotation of a ScaleCUA mobile trajectory with an unnecessary action (Step 3, highlighted). The teacher outputs three subtasks and flags the redundant scroll action.}
    \label{fig:teacher_unnecessary}
\end{figure}

\begin{figure}[h]
    \centering
    \includegraphics[width=\linewidth]{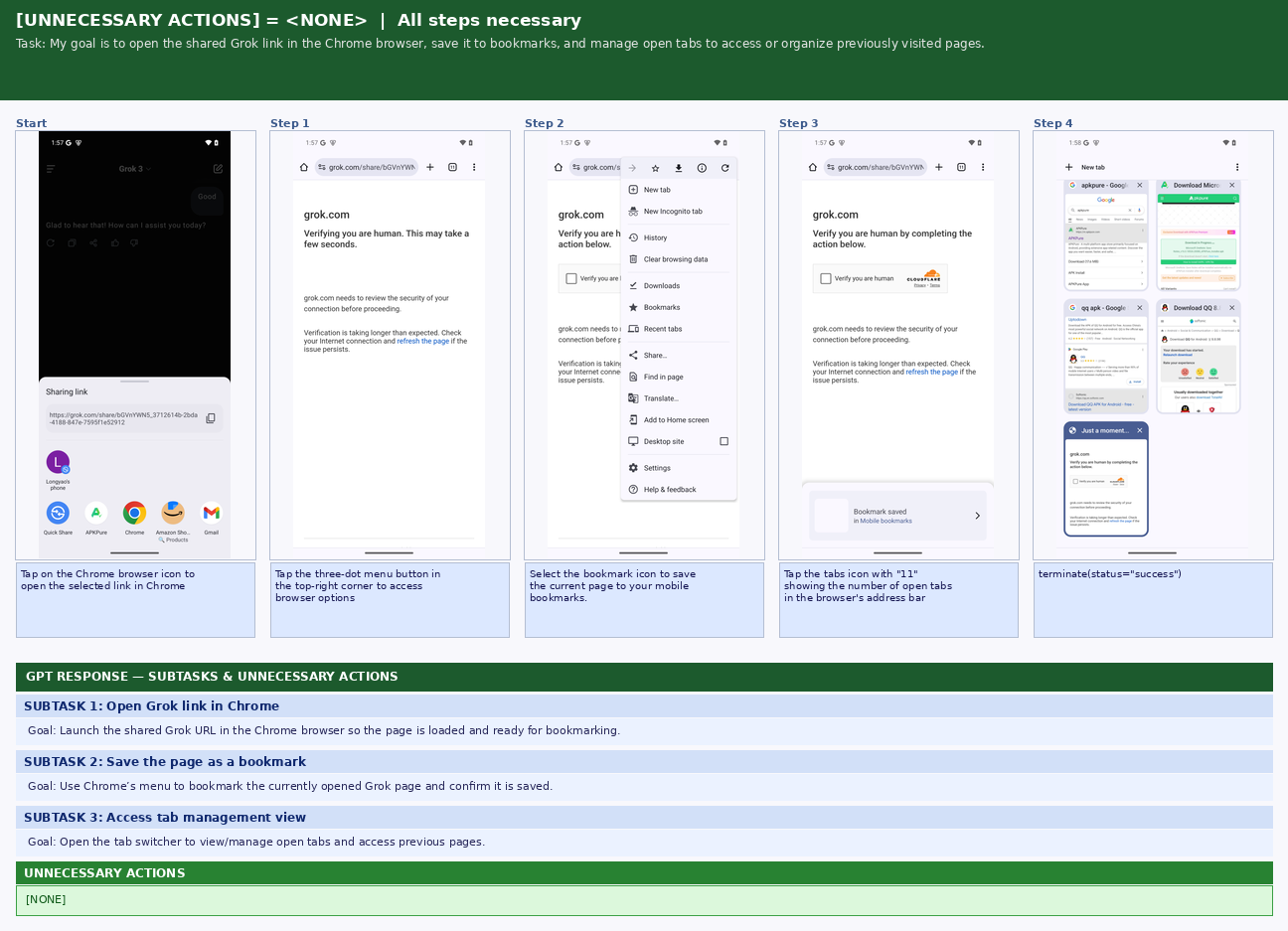}
    \caption{Teacher annotation of a ScaleCUA mobile trajectory in which all actions are necessary; the teacher outputs three subtasks and \texttt{[NONE]} for unnecessary actions.}
    \label{fig:teacher_none}
\end{figure}

\begin{figure}[h]
    \centering
    \includegraphics[width=\linewidth]{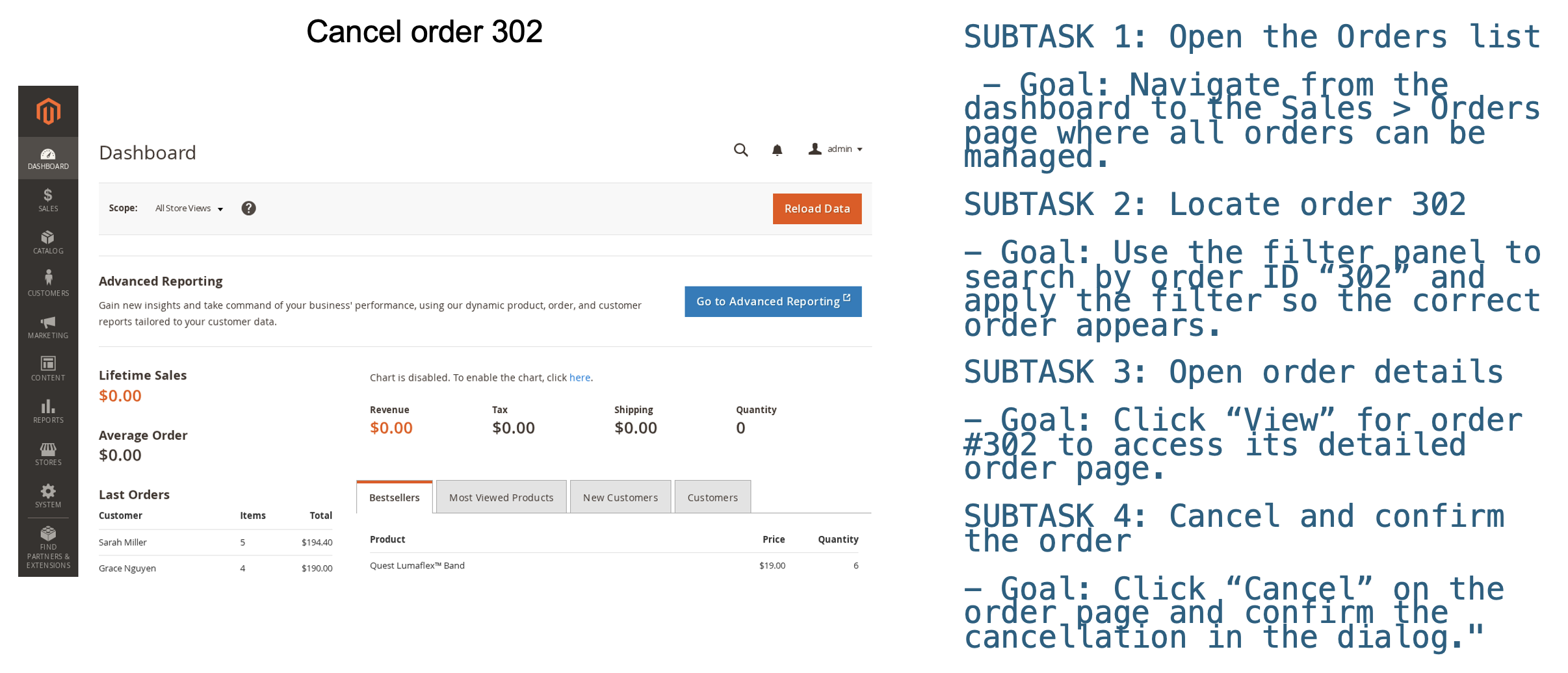}
    \caption{Decomposition generated by the JHiMA-WebRL decomposer for the ShoppingAdmin evaluation task ``Cancel order 302'' from the task instruction and starting screenshot.}
    \label{fig:generated_decomp}
\end{figure}

\begin{figure}[h]
    \centering
    \includegraphics[width=\linewidth]{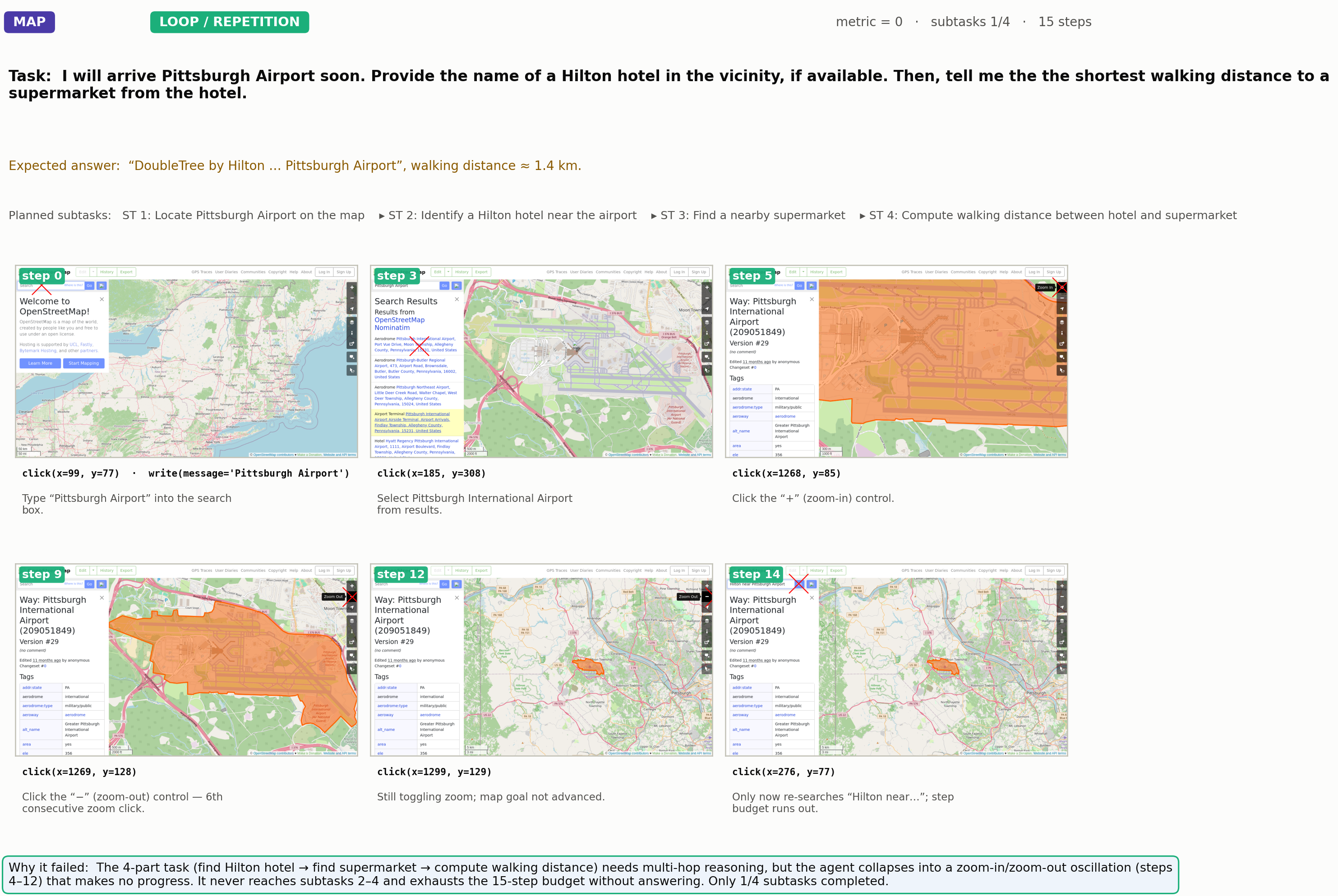}
    \caption{Failure of JHiMA-WebRL on a Map task: repeated irrelevant actions exhaust the 15-step budget.}
    \label{fig:map_failure}
\end{figure}

Figures~\ref{fig:teacher_unnecessary} and~\ref{fig:teacher_none} show teacher-generated annotations for two ScaleCUA trajectories, one containing an unnecessary action and one in which all actions are necessary. Figure~\ref{fig:generated_decomp} shows a decomposition produced by our decomposer from the task instruction and starting screenshot of a WebArena ShoppingAdmin task. Figure~\ref{fig:map_failure} shows a failure case for a Map task.

\end{document}